%% file: main.tex
\documentclass{article}
\usepackage{spconf,amsmath,graphicx,dsfont,amsfonts,url,subfigure,wrapfig}

\title{Large-Scale Deep Learning on the YFCC100M Dataset}
\name{Karl Ni, Roger Pearce, Eric Wang, Kofi Boakye, Brian Van Essen, Damian Borth, Barry Chen\thanks{This work was performed under the auspices of the U.S. Department of Energy by Lawrence Livermore National Laboratory in part under Contract W-7405-Eng-48 and in part under Contract DE-AC52-07NA27344.}}
\address{Lawrence Livernmore National Laboratory \\
         Computational Engineering Division \\
         7000 East Avenue, Livermore, CA 94550}
\begin{document}
\ninept
\maketitle
\begin{abstract}
We present a work-in-progress snapshot of learning with a 15 billion parameter deep learning network on HPC architectures applied to the largest publicly available natural image and video dataset released to-date.  Recent advancements in unsupervised deep neural networks suggest that scaling up such networks in both model and training dataset size can yield significant improvements in the learning of concepts at the highest layers.  We train our three-layer deep neural network on the Yahoo! Flickr Creative Commons 100M dataset.  The dataset comprises approximately $99.2$ million images and $800,000$ user-created videos from Yahoo's Flickr image and video sharing platform.  Training of our network takes eight days on 98 GPU nodes at the High Performance Computing Center at Lawrence Livermore National Laboratory.  Encouraging preliminary results and future research directions are presented and discussed.
\end{abstract}

\section{Introduction}
\input{introduction}

\section{Overview of the YFCC100M Dataset}\label{sec:dataset}
\input{yfccdata}

\section{Analysis with Large Scale Neural Networks}\label{sec:network}
\input{architecture}

% \section{Large Neural Networks on an HPC Architecture}\label{sec:hpc}

\section{Preliminary Results}\label{sec:results}
\input{results}
\input{summary}

\section{Acknowledgments}
We would like to thank Adam Coates, Brody Huval and Andrew Ng for providing their COTS HPC Deep Learning software and helpful advice.  This work was performed under the auspices of the U.S. Department of Energy by Lawrence Livermore National Laboratory under Contract DE-AC52-07NA27344.

% \bibliography{nipsworkshop.bib}
\newpage
\bibliographystyle{IEEEbib}
\bibliography{nipsworkshop}

\end{document}

%% file: introduction.tex
The field of deep learning via stacked neural networks has received renewed interest in the last decade \cite{krizhevsky12, coates13, le12}. Neural networks have been shown to perform well in a wide variety of tasks, including text analysis \cite{mikolov_1}, speech recognition \cite{Abdel-Hamid13, DNN_ASR2012,BourlardAndMorgan}, various classification tasks \cite{Ciresan11,Reby97}, and most notably unsupervised and supervised feature learning on natural imagery \cite{krizhevsky12, coates13, le12}.

Deep neural networks applied to natural images have demonstrated state-of-the-art performance in supervised object recognition tasks \cite{uetz09,krizhevsky12} as well as unsupervised neural networks \cite{coates13,le12}.  The classical approach to training neural networks for computer vision is via a large dataset of labeled data.  However, sufficiently large and accurately labeled data is difficult and expensive to acquire.  Motivated by this, \cite{le12} explored the application of deep neural networks in unsupervised deep learning and discovered that sufficiently large deep networks are capable of learning highly complex concept level features at the top level without labels.

Spurred by this advancement, \cite{coates13} set out to construct very large networks on the order of $10^9$ to $10^{10}$ parameters.  A key advancement was the  highly efficient multi-GPU architecture of their model.  \cite{coates13} employed a high degree of model parallelism and was able to process 10 million YouTube thumbnails in a few days processing time on a medium sized cluster.    A notable result was the unsupervised learning of various faces, including those of humans and cats.  Ultimately, improved feature learning at larger scales can improve downstream capabilities such as scene or object classification, additional unsupervised learning (\emph{i.e.} via topic modeling \cite{Cao07} or natural language processing algorithms \cite{Socher13}).

In collaboration with the authors of \cite{coates13}, we have scaled a similar model and architecture to over 15 billion parameters on the Lawrence Livermore National Laboratory's (LLNL) Edge High Performance Computing (HPC) system.  Our long-term goal is two-fold: (1) explore at-the-limit performance of massive networks ($>10$ billion parameters) and (2) train on and analyze datasets on the order of 100 million images.  

As the number of network parameters grow, datasets need to be scaled accordingly to avoid overfitting the models.  We take advantage of a brand-new dataset released jointly by Yahoo!, LLNL and the International Computer Science Institute (ICSI) called the Yahoo! Flickr Creative Commons 100M (YFCC100M) dataset10.  The dataset is, to the authors' knowledge, the largest single publicly available image and video dataset ever published.  In addition to the raw images and video, the YFCC100M also contains metadata for each entry including locations, camera types, keywords, titles, etc.  Although beyond the scope of this paper, this rich associated meta-data potentially offers researchers additional avenues of semantic multi-modality learning to explore.

Working with the large-scale datasets, models and computing architectures considered in this paper presents several daunting engineering challenges. For example, the significantly greater number of GPUs and compute nodes used in our system versus \cite{coates13} creates communication issues in MPI. 
%while LLNL computational resources can include large amounts of per-node memory, themodel is distributed among 98 GPUs, which have significantly less on-board memory. This creates communication issues in MPI. 
%In addition, because of the way the model is distributed on our system, there are certain parameter combinations that are computationally infeasible. 
In addition, a typical model takes up over 40~GB of memory, making simple offline analysis tasks such as visualization challenging. Various network architectures were tested, balancing performance and computational constraints, before we arrived at our current model.  Finally, as in \cite{coates13}, data throughput presents a bottleneck to model training. We present a novel pipeline approach  to address this problem. 

%meaning that the full network may not be viewed with ordinary desktop machines. [FILL IN SOME HIGH LEVEL CHALLENGES TO MAKE US LOOK LIKE BAMFs.  Some examples include: the 64 GPU issue, memory issues, data issues, etc. - BRIAN, ROGER, KOFI].

% this paper describes ongoing work in applying extremely large ($\approx$ 10 billion parameters) neural networks to the new Yahoo FlickR Creative Commons 100M-14G (YFCC-100M) dataset on the extreme large HPC infrastructure at the Lawrence Livermore National Laboratory (LLNL). The YFCC-100M dataset boasts being the largest source of creative commons image and video data for academic and research purposes, opening the door to large scale multi-modal exploration. YFCC-100M metadata contains captions, locations, camera types, keywords, titles, etc., which also enables semantic exploration and learning, which we also address in the paper.

The rest of this paper is organized as follows.  In Section \ref{sec:dataset} we give a brief overview of the YFCC100M dataset.  The network architecture and computational framework being employed is described in Section \ref{sec:network}.  We present preliminary results and visualizations of our network in Section \ref{sec:results}.  Finally, we summarize and discuss future research directions in Section \ref{sec:summary}.

%% file: yfccdata.tex
In late June 2014, Yahoo! released the Yahoo! Flickr Creative Commons dataset (YFCC100M). This dataset consists of 100 million Flickr user-uploaded images and videos (99,206,564 images and 793,436 videos) along with their corresponding metadata including title, description, camera type, tags, and geotags when available.  All of the data is under Creative Commons licensing and is freely provided to scientists for the advancement of multimedia research~\footnote{Available at http://research.yahoo.com/Academic\_Relations}. In addition to the raw images, videos, and metadata, Yahoo! in collaboration with the ICSI and LLNL will be computing and providing standard computer vision and audio features using LLNL's supercomputing resources.

Wang et al. \cite{WangVL14} have used YFCC100M data to build systems that associate images with more natural annotations like those found in user-generated captions. Others are interested in using the YFCC100M imagery and audio to geolocate where the photo or video was taken \cite{im2gps}. In fact, the 2014 MediaEval Placing Task is using YFCC100M as the source of benchmark data \cite{geoMM}. We are interested in using YFCC100M as our sandbox dataset for learning image features using massive unsupervised neural networks, repeating the experiment by \cite{le12} on an order of magnitude more data and neural network parameters.  In particular, we want to see what other ``grandmother neurons'' \cite{le12} our network would automatically learn from YFCC100M.

\begin{table}[ht]
{\tiny
  \caption {Top 60 Tags in YFCC100M Images} \label{tab:toptags}
  \begin{center}
    \begin{tabular}{| c c c c c c|}
      \hline
square & iphoneography & square format & instagram app & california & travel \\
nikon & usa & canon & london & japan & france \\
nature & art & music & europe & beach & united states \\
england & wedding & italy & new york & canada & city \\
vacation & germany & party & park & water & people \\
uk & spain & architecture & summer & festival & nyc \\
taiwan & paris & san francisco & australia & winter & sky \\
snow & concert & night & family & china & museum \\
food & street & live & washington & landscape & flower \\
sunset & photo & flowers & holiday & trip & photography \\
      \hline
    \end{tabular}
  \end{center}
}
\end{table}

\begin{figure*}[ht]
\centering
%\framebox[4.0in]{$\;$}
% \fbox{\rule[-.5cm]{0cm}{4cm} \rule[-.5cm]{4cm}{0cm}}
\begin{subfigure}{\includegraphics[scale=0.15]{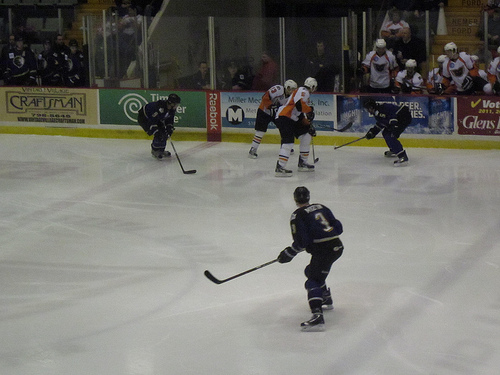}}
\end{subfigure}
\begin{subfigure}{\includegraphics[scale=0.15]{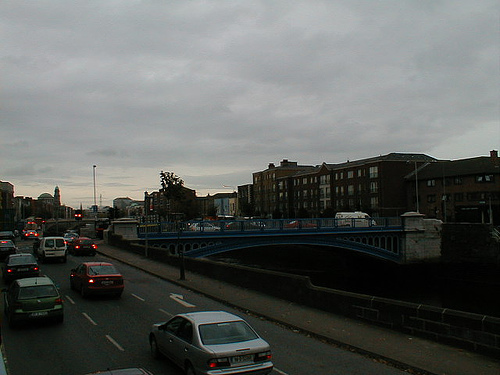}}
\end{subfigure}
\begin{subfigure}{\includegraphics[scale=.26]{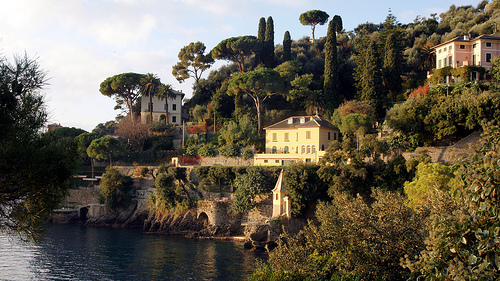}}
\end{subfigure}
\begin{subfigure}{\includegraphics[scale=6.75]{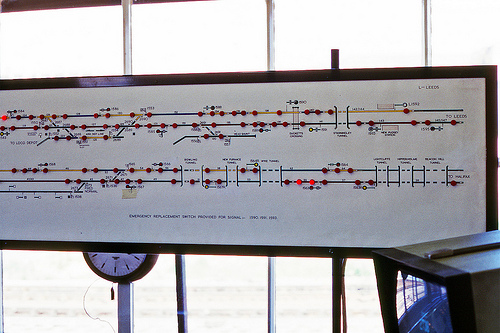}}
\end{subfigure}
\begin{subfigure}{\includegraphics[scale=0.15]{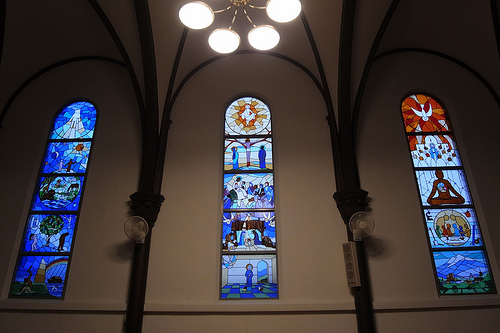}}
\end{subfigure}
\begin{subfigure}{\includegraphics[scale=0.42]{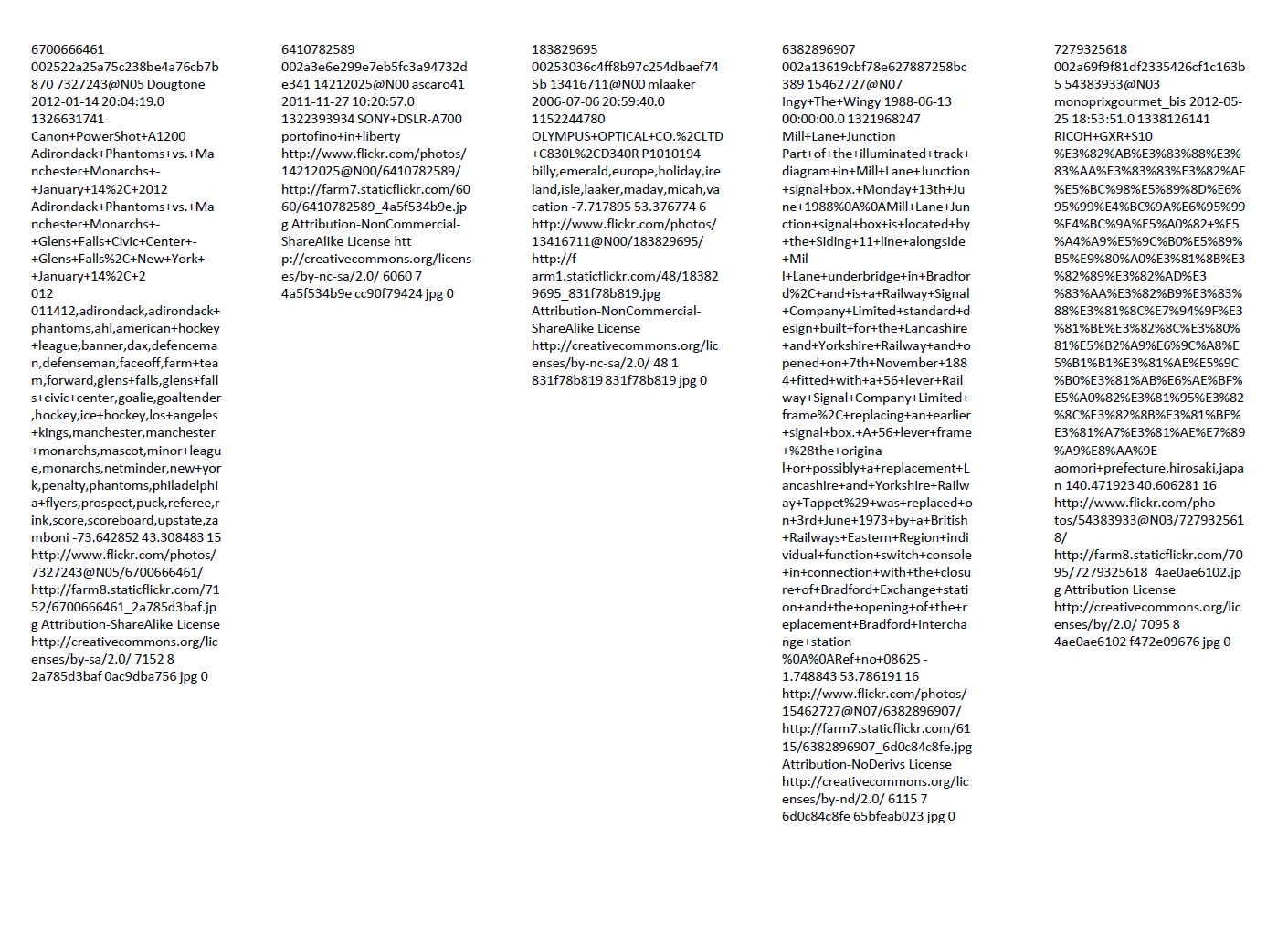}}
\end{subfigure}
 \vspace{-15mm}
\caption{Examples of YFCC Data, and the associated metadata. Photo credits to Yahoo! users ``Dougtone'', ``ascaro41'', ``mlaaker'', ``Ingy The Wingy'', ``monoprixgourmet\_bis''.}
\label{exampleyfcc}
\end{figure*}

The 99,206,564 images were created and posted by 578,268 different Flickr users. 76\%, 20\%, and 4\% of the images have titles, auto-titles, or no titles, respectively.  The average number of words per title is 3.08. 32\% of the images have descriptions with an average of 22.52 words per description.  Finally, 69\% of the images have on average 7.07 tags per image. The top 60 tags are shown in Table~\ref{tab:toptags}.  In Fig.~\ref{exampleyfcc} we show example images and associated meta-data for several YFCC100M images.
 
 % \vspace{-1mm}

%% file: architecture.tex
\subsection{Network Architecture}

For the large set of image data, we employed a three-layer, large-scale deep neural network with a reconstruction independent component analysis (RICA) cost function, 

\begin{equation}
\label{RICA}
\begin{split}
\min_{W,\alpha,b} & \sum_i \left\| W^T (\alpha W x^{(i)}) + b - x^{(i)} \right\|_2^2 + \lambda \sqrt{ (\alpha W x^{(i)})^2 } \\
 & \text{ subject to } \| W^{(k)} \|_2 = 1, \forall k,\nonumber
\end{split}
\end{equation}

where as in~\cite{coates13}, $W$ is a weighting matrix, $\alpha$ is a scaling value and $x^{(i)}$ are the data points at the beginning of each layer. In addition, we introduce an offset, $b$, for increased model flexibility. The parameter $\lambda$ controls the relative sparsity, and is set to 0.1 at the first two layers and 0.01 at the final layer.  Unlike \cite{coates13}, we do not presently include a pooling layer, as we believe the scale of the network and training data allows a similar translational invariance to be automatically learned. A particular advantage conferred by the RICA construction is that the sparseness term $\lambda \sqrt{ (\alpha W x^{(i)})^2 }$ can be computed in-situ with the rest of the model parameters.  This is in contrast to the conventional sparse autoencoder construction that requires a second pass through the data to compute a sparseness-specific gradient contribution.

\begin{figure}[ht]
\begin{center}
%\framebox[4.0in]{$\;$}
% \fbox{\rule[-.5cm]{0cm}{4cm} \rule[-.5cm]{4cm}{0cm}}
\includegraphics[scale=0.3]{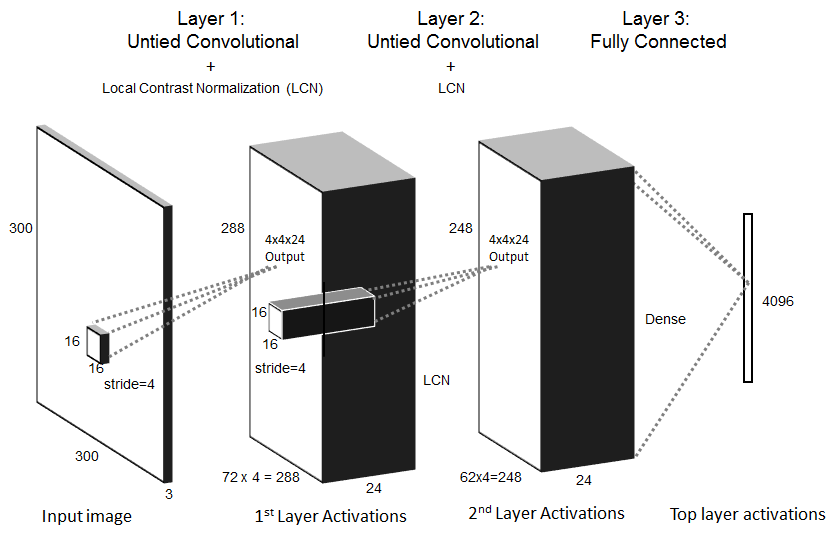}
\end{center}
\caption{Network topology of large scale, trained network.  Approximately 15 billion parameters}
\label{topology}
\end{figure}

Fig.~\ref{topology} illustrates the structure of our network. The three layers are composed of two untied convolutional layers, and a third fully-connected layer. The first convolutional layer utilizes 5184 filters \footnote{Arranged in a $72\times72$ grid} of input size $16\times16\times3$ with stride $4$ and output size $4\times4\times24$. The second layer takes $16$ spatially contiguous \footnote{Arranged in a $4\times4$ grid} $4\times4\times24$ outputs of the first layer and connects them fully to a $4\times4\times24$ output.  The stride length of the second layer is 4.  The third layer is dense, and fully connects the $62\times62\times24$ outputs of the second layer to 4096 top-level neurons.  The total number of parameters trained is 15 billion. After each layer, local contrast normalization (LCN) is applied prior to continuing onto the next layer. Though no pooling is applied, the window sizes at the next layer are large enough to incorporate spatial information from neighboring blocks. 
%
% \begin{wrapfigure*}{r}{0.5\textwidth}
% \begin{center}
% %\framebox[4.0in]{$\;$}
% % \fbox{\rule[-.5cm]{0cm}{4cm} \rule[-.5cm]{4cm}{0cm}}
% \includegraphics{topology.png}
% \end{center}
% \caption{Network topology of large scale, trained network.  Approximately 15 billion parameters}
% \label{topology}
% \end{wrapfigure*}
%
\begin{figure}[ht]
\begin{center}
%\framebox[4.0in]{$\;$}
% \fbox{\rule[-.5cm]{0cm}{4cm} \rule[-.5cm]{4cm}{0cm}}
\includegraphics[scale=0.3]{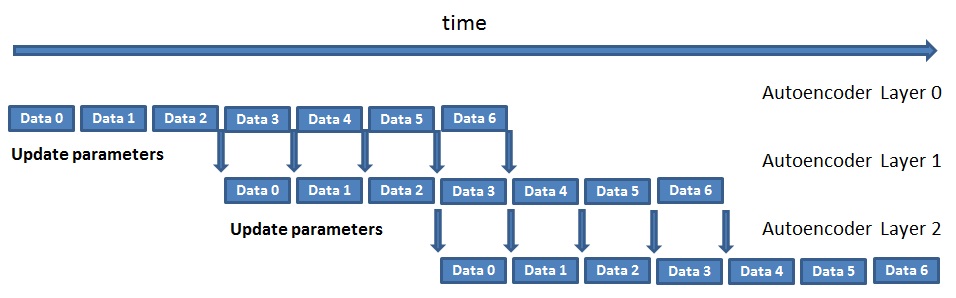}
\end{center}
\caption{Pipeline for semi-parallel training of sparse autoencoders from a single data source}
\label{onlinetrain}
\end{figure}

Training data is arranged into 99,207 data blocks of 960 images. Each data block consists of 5 mini-batches, where each mini-batch contains 192 images. Due to the scale of the data, the proposed algorithm reduces training time by employing a pipeline technique where the next layer begins training before the previous layer has finished. Analogous to the example shown in Fig.~\ref{onlinetrain}, after a layer $L$ has trained an initial set of data blocks (in our case, 1000), the next layer, $L+1$, starts training. To accomplish this, two instances of the layer $L$ are run simultaneously: one which continues training and one that uses up-to-date parameters to forward propagate data from Block 0 to the layer $L+1$. The parameters of the forward-propagating layer $L$ instance are periodically synchronized with the layer $L$ instance that continued training.  We observed that our model was not sensitive to the choice of synchronization frequency.  As a rule of thumb, we wait to train layer $L+1$ until the objective of layer $L$ stabilizes, which typically occurs after approximately one million images.

\subsection{HPC Architecture}
\input{hpc}

%% file: hpc.tex
To train the neural network at scale, we used 98 nodes of the Edge HPC cluster at Lawrence Livermore National Laboratory.  The Edge cluster consists of 206 nodes with 12 core Intel Xeon EP X5660 running at 2.8~GHz.  Each node has 96~GB of DRAM and a Tesla M2050 (Fermi) NVIDIA GPU with 3~GB of GDDR5. The training algorithm is model parallel as described in \cite{coates13}, with the nodes and GPUs processing each mini-batch across the system and distributing the model across the GPUs. Communication was provided by MPI over Mellanox QDR Infiniband cards.  The GPU accelerators were used with CUDA 5.5 and MPI-direct communication and the operating system was a 2.6.32 kernel RHEL 6 derivative.

The dataset was stored in a Lustre file system with a peak bandwidth of 10~GB/s. Each mini-batch was copied from Lustre into memory and then streamed into the GPU's memory.  Each GPU is responsible for computing its section of the model parameters for the current mini-batch.  Communication within the algorithm
occurs when a layer's input (or output) field spans multiple GPUs. The communication is handled by a distributed array data structure (using MPI) within the training algorithm.  Global communication is minimized by using untied local receptive fields, and allowing receptive fields to be trained independently.

%% file: results.tex
\begin{figure}[ht!]
\centering
%\framebox[4.0in]{$\;$}
% \fbox{\rule[-.5cm]{0cm}{4cm} \rule[-.5cm]{4cm}{0cm}}
\begin{subfigure}{\includegraphics[scale=0.152]{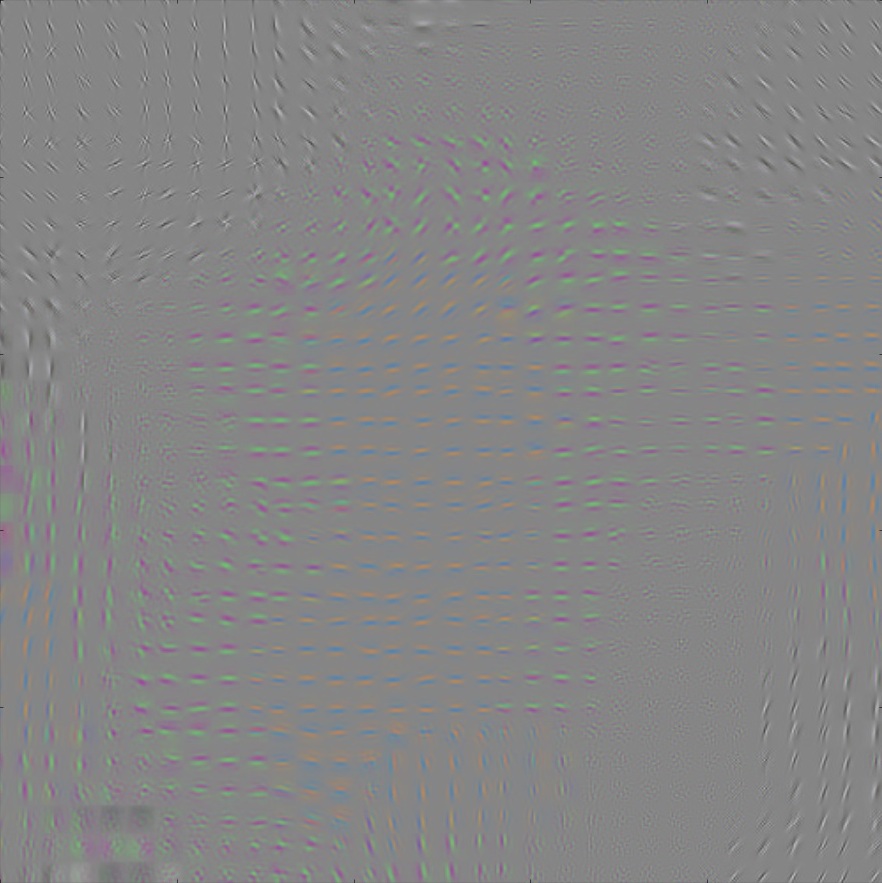}}
\end{subfigure}
\begin{subfigure}{\includegraphics[scale=0.154]{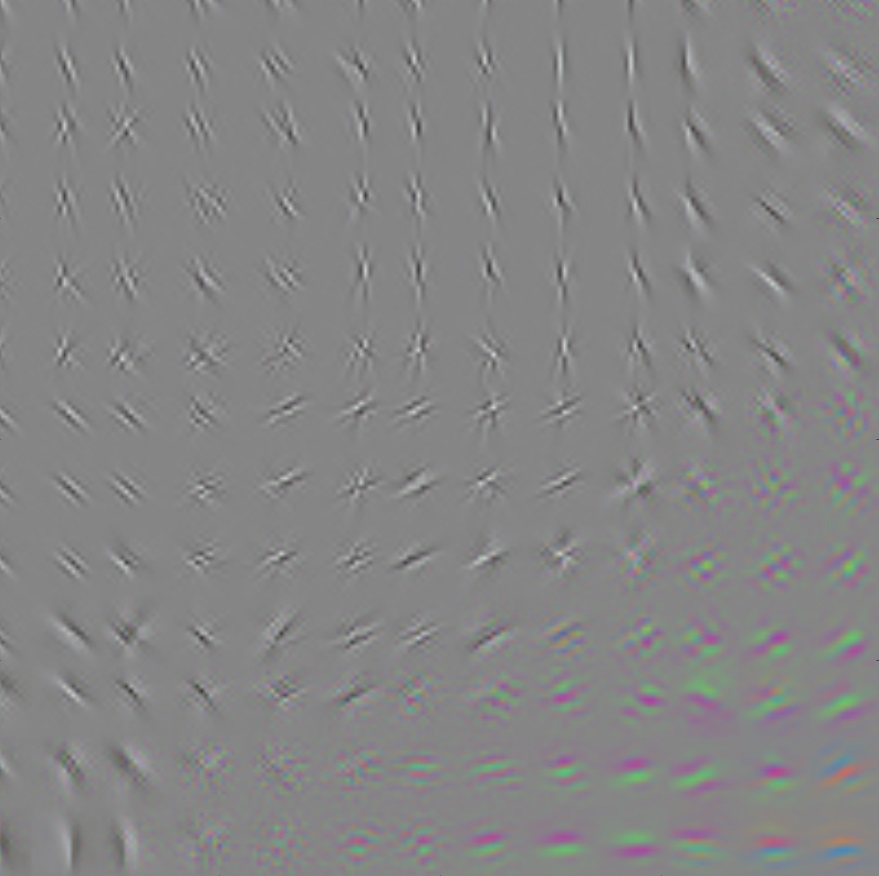}}
\end{subfigure}
\caption{Visualization of a selection of typical first layer weights.  The right figure is a zoomed-in crop of the left.}
\label{firstlayerweights}
\end{figure}

% \begin{wrapfigure}{r}{0.5\textwidth}
% \centering
% %\framebox[4.0in]{$\;$}
% % \fbox{\rule[-.5cm]{0cm}{4cm} \rule[-.5cm]{4cm}{0cm}}
% \begin{subfigure}{\includegraphics[scale=0.278]{filters_l0_b.jpg}}
% \end{subfigure}
% \begin{subfigure}{\includegraphics[scale=0.28]{filters_l0_c.jpg}}
% \end{subfigure}
% \caption{Visualization of a selection of typical first layer weights.  The right figure is a zoomed-in crop of the left.}
% \label{firstlayerweights}
% \end{wrapfigure}

% \begin{figure}[ht!]
% \centering
% \begin{tabular}{cc}
% \hspace{-5mm}{\includegraphics[scale=0.145]{neuron_1030.jpg}} & \hspace{-9mm}{\includegraphics[scale=0.145]{neuron_1104.jpg}} \\
% \hspace{-5mm}{\includegraphics[scale=0.145]{neuron_1296.jpg}} & \hspace{-9mm}{\includegraphics[scale=0.145]{neuron_1460.jpg}} \\
% \hspace{-5mm}{\includegraphics[scale=0.145]{neuron_1506.jpg}} & \hspace{-9mm}{\includegraphics[scale=0.145]{neuron_1905.jpg}} \\
% \hspace{-5mm}{\includegraphics[scale=0.145]{neuron_3791.jpg}} & \hspace{-9mm}{\includegraphics[scale=0.145]{neuron_2289.jpg}} 
% \end{tabular}
% \caption{Top 5 Stimuli of example layer 3 neurons.  Images have been whitened.}
% \label{neuronexamples}
% \end{figure}

\begin{figure}[!ht]
\centering
\begin{subfigure}{\includegraphics[scale=0.15]{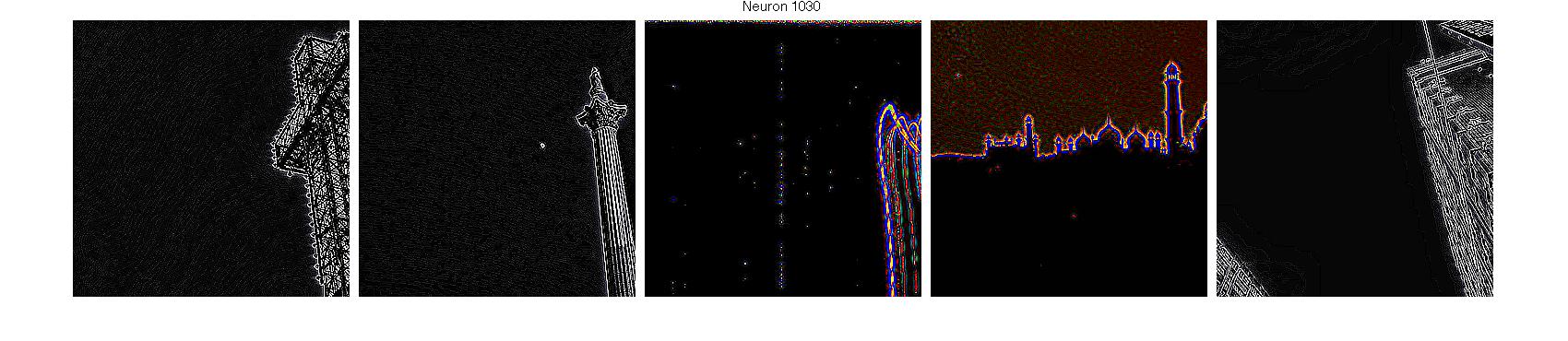}}
\end{subfigure}
% \end{figure}
%
% \begin{figure}[!ht]
% \centering
\begin{subfigure}{\includegraphics[scale=0.15]{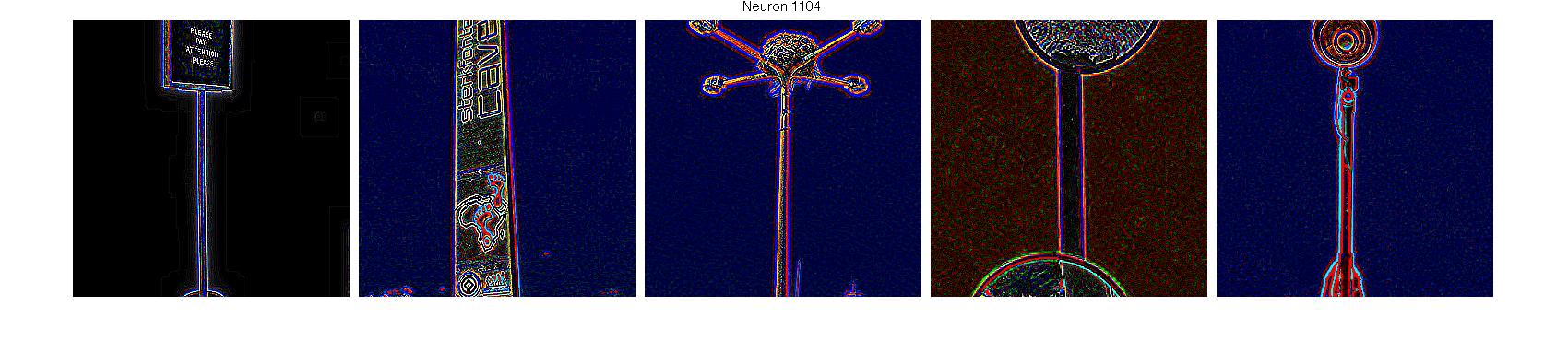}}
\end{subfigure}
\begin{subfigure}{\includegraphics[scale=0.15]{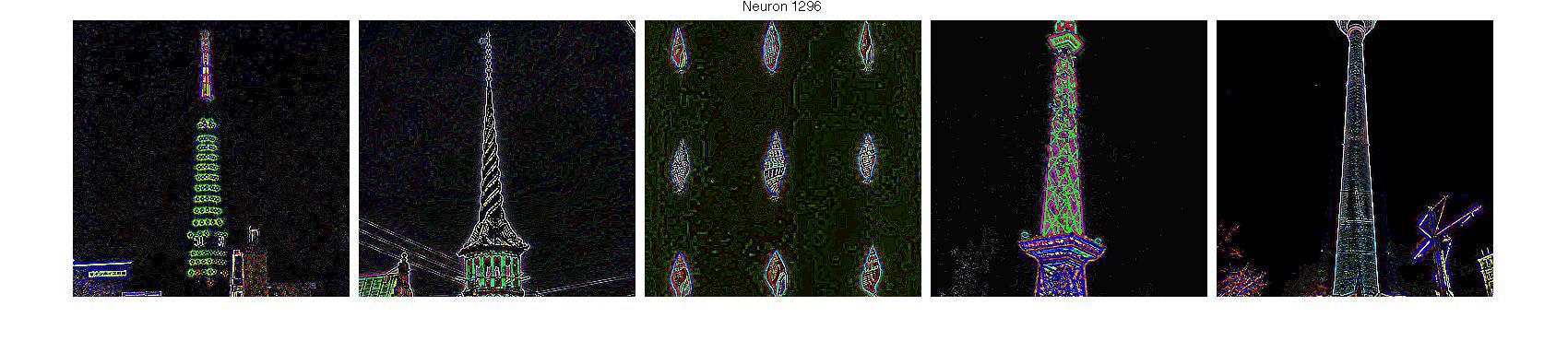}} 
\end{subfigure}
\begin{subfigure}{\includegraphics[scale=0.15]{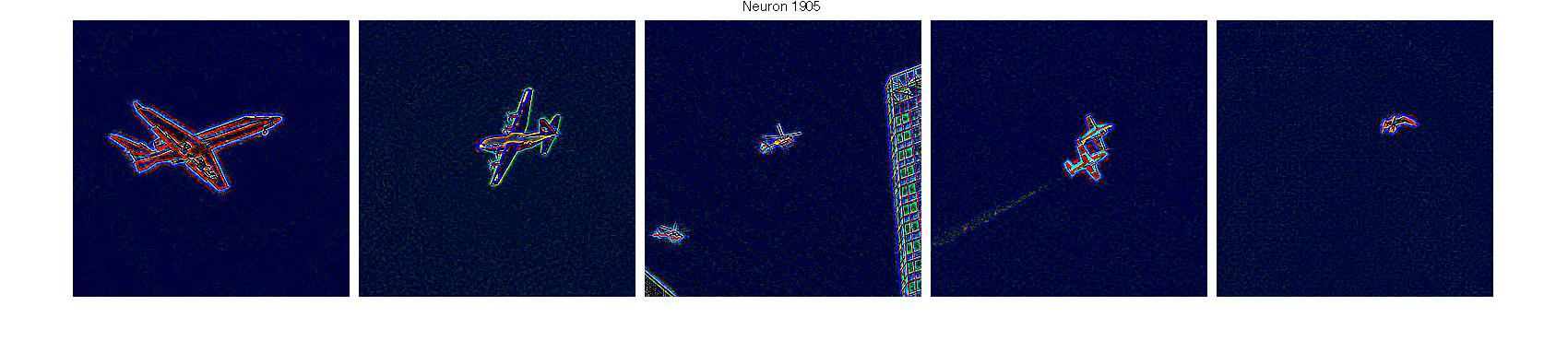}}
\end{subfigure}
\begin{subfigure}{\includegraphics[scale=0.15]{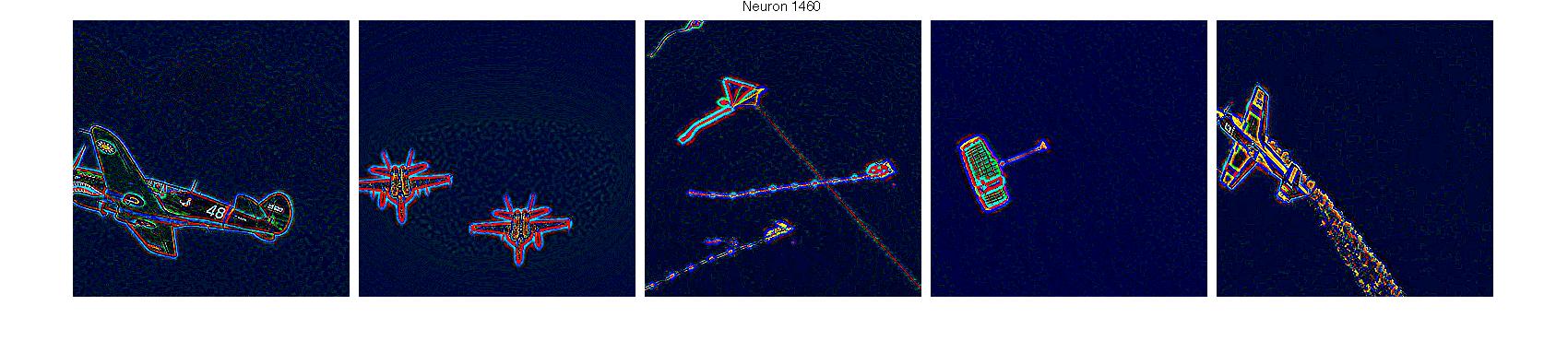}} 
\end{subfigure}
\begin{subfigure}{\includegraphics[scale=0.15]{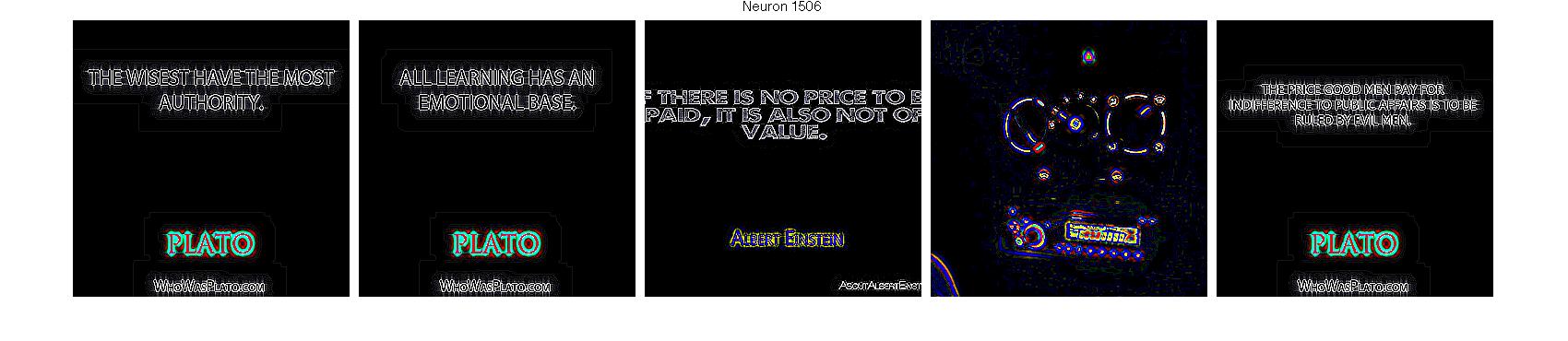}}
\end{subfigure}
\begin{subfigure}{\includegraphics[scale=0.15]{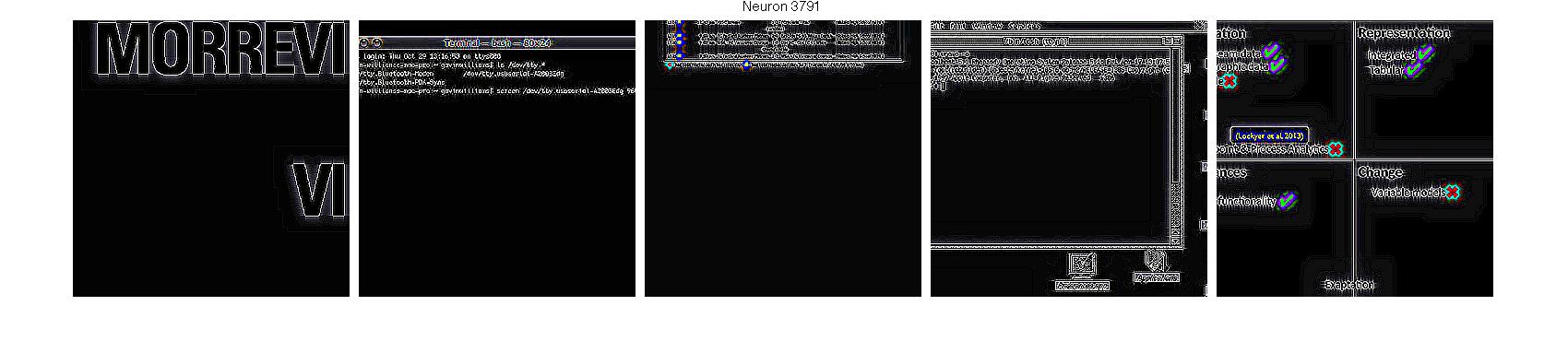}} 
\end{subfigure}
\begin{subfigure}{\includegraphics[scale=0.15]{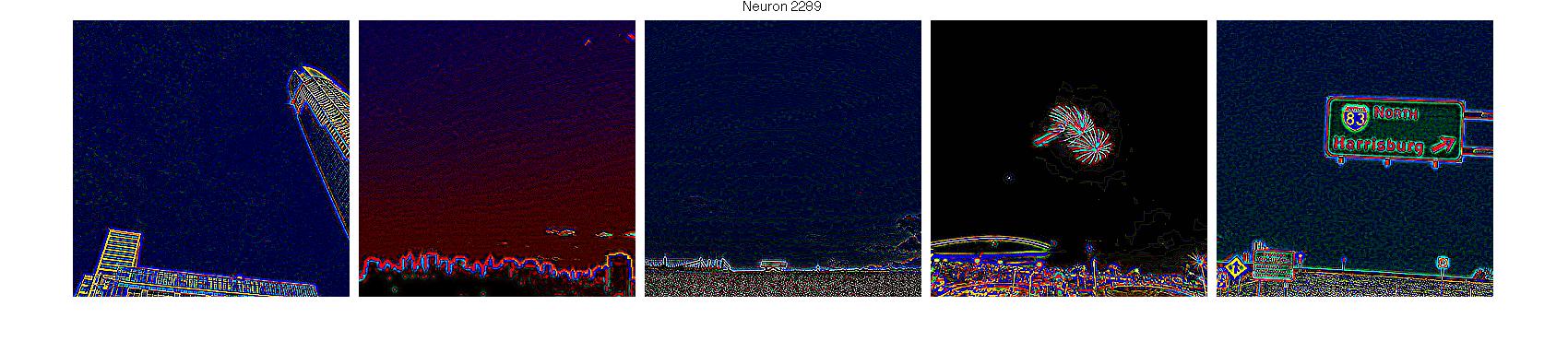}} 
\end{subfigure}
\caption{Top-5 stimuli of example layer 3 neurons.  Images have been whitened.}
\label{neuronexamples}
\end{figure}

We trained the network using all images from the YFCC100M dataset.  Images were preprocessed as in \cite{coates13}, and subsequently resized to 300 x 300 pixels by first centering, then scaling the smallest dimension to 300 pixels, and finally cropping. After training all three layers, we forward propagated 2 million images through the network in order to obtain activation values for visualization.  Note that in this paper, the test set is significantly noisier than the benchmark Labeled Faces In the Wild \cite{huang07} and ImageNet \cite{deng09} datasets considered in previous works such as \cite{le12}. 

% In Fig.~\ref{firstlayerweights} we visualize some typical first layer weights.  As expected, they are trained to capture various types of edges and separate into color and texture focused neurons. 
In Fig.~\ref{neuronexamples}, we show the top 5 stimuli for some example neurons.  We observe that our network is capable of learning significant structure, identifying buildings, aircraft, text, cityscapes, and tower-like buildings, among many others.  The network seems to cue in on distinctive textures such as the edges of text, sides of buildings and the sharp edge of airplanes against the smooth gradation of the sky. Moreover, the network seems to activate on large-scale structures within an image rather than local features.  We believe that a significant contributor to our networks' performance is due to its large size being able to capture complex concepts.  

While our results are encouraging, we believe that significant improvements in learning can be achieved through improved network architecture and increased depth.  As was demonstrated in \cite{krizhevsky12}, network architecture has a significant impact on the performance of deep networks.  While the networks described in \cite{le12} were able to learn complex features in just three layers, our results suggest that extremely large datasets such as the YFCC100M can support (and possibly benefit from) deeper networks with improved high-level concept learning. 

%% file: summary.tex
\section{Summary and Future Work}\label{sec:summary}
The results discussed in this paper present a snapshot of the work in progress at Lawrence Livermore National Laboratory in scaling up deep neural networks.  Such networks offer enormous potential to researchers in both supervised and unsupervised computer vision tasks, from object recognition and classification to unsupervised feature extraction.

To date, we see highly encouraging results from training our large 15 billion parameter three-layer neural network on the YFCC100M dataset in an unsupervised manner.  The results suggest that the network is capable of learning highly complex concepts such as cityscapes, aircraft, buildings, and text, all without labels or other guidance.  That this structure is visible upon examination is made all the more remarkable due to the noisiness of our test set (taken at random from the YFCC100M dataset itself).

Future work on our networks will focus on two main thrusts: (1) improve the high-level concept learning by increasing the depth of our network, and (2) scaling our network's width in the middle layers.  On the first thrust, we aim for improved high-level summarization and scene understanding.  Challenges on this front include careful tuning of parameters to combat the ``vanishing gradient" problem and design of the connectivity structure of the higher-level layers to maximize learning.  On the second thrust, our challenges are primarily engineering focused.  Memory and message passing constraints become a serious concern, even on the large HPC systems fielded by LLNL.  As we move beyond our current large neural network, we plan to explore the use of memory hierarchies for staging intermediate/input data to minimize the amount of node-to-node communication, enabling the efficient training and analysis of even larger networks.

%% file: main.bbl
\begin{thebibliography}{10}

\bibitem{krizhevsky12}
A.~Krizhevsky, I.~Sutskever, and G.~Hinton,
\newblock ``Imagenet classification with deep convolutional neural networks,''
\newblock in {\em Advances in neural information processing systems}, 2012.

\bibitem{coates13}
A.~Coates, B.~Huval, T.~Wang, D.~J. Wu, A.~Y. Ng, and B.~Catanzaro.,
\newblock ``Deep learning with cots hpc,''
\newblock in {\em International Conference on Machine Learning}, 2013.

\bibitem{le12}
Q.~V. Le, M.~Ranzato, R.~Monga, M.~Devin, K.~Chen, G.~S. Corrado, J.~Dean, and
  A.~Y. Ng,
\newblock ``Building high-level features using large scale unsupervised
  learning,''
\newblock in {\em International Conference on Machine Learning}, 2012.

\bibitem{mikolov_1}
T.~Mikolov, K.~Chen, G.~Corrado, and J.~Dean.,
\newblock ``Efficient estimation of word representations in vector space,''
\newblock in {\em Proceedings of Workshop at ICLR}, 2013.

\bibitem{Abdel-Hamid13}
O.~Abdel-Hamid, L.~Deng, and D.~Yu,
\newblock ``Exploring convolutional neural network structures and optimization
  techniques for speech recognition,''
\newblock in {\em Interspeech 2013}, 2013.

\bibitem{DNN_ASR2012}
G.~Hinton, L.~Deng, D.~Yu, A.-R. Mohamed, N.~Jaitly, A.~Senior, V.~Vanhoucke,
  P.~Nguyen, T.~Sainath, G.~Dahl, and B.~Kingsbury,
\newblock ``Deep neural networks for acoustic modeling in speech recognition,''
\newblock {\em IEEE Signal Processing Magazine}, vol. 29, no. 6, pp. 82--97,
  November 2012.

\bibitem{BourlardAndMorgan}
H.~Bourlard and N.~Morgan,
\newblock {\em Connectionist Speech Recognition: A Hybrid Approach},
\newblock Kluwer Academic Publishers, 1993.

\bibitem{Ciresan11}
D.~Claudiu Ciresan, U.~Meier, L.~M. Gambardella, and J.~Schmidhuber,
\newblock ``Convolutional neural network committees for handwritten character
  classification,''
\newblock in {\em International Conference on Document Analysis and
  Recognition}, 2011.

\bibitem{Reby97}
D.~Reby, S.~Lek, I.~Dimopoulos, J.~Joachim, J.~Lauga, and S.~Aulagnier,
\newblock ``Artificial neural networks as a classification method in the
  behavioural sciences,''
\newblock {\em Behavioural Processes}, vol. 40, pp. 35–43, 1997.

\bibitem{uetz09}
R~Uetz and S.~Behnke,
\newblock ``Large-scale object recognition with cuda-accelerated hierarchical
  neural networks,''
\newblock in {\em IEEE International Conference on Intelligent Computing and
  Intelligent Systems}, 2009.

\bibitem{Cao07}
L.~Cao and L.~Fei-Fei,
\newblock ``Spatially coherent latent topic model for concurrent object
  segmentation and classification,''
\newblock in {\em Proceedings of International Conference on Computer vision},
  2007.

\bibitem{Socher13}
{R. Socher and M. Ganjoo and C. D. Manning and A. Y. Ng},
\newblock ``{Zero Shot Learning Through Cross-Modal Transfer},''
\newblock in {\em {Advances in Neural Information Processing Systems 26}}.
  2013.

\bibitem{WangVL14}
J.~K. Wang, F.~Yan, A.~Aker, and R.~Gaizauskas,
\newblock ``A poodle or a dog? {E}valuating automatic image annotation using
  human descriptions at different levels of granularity,''
\newblock in {\em Proceedings of the Workshop on Vision and Language}, 2014.

\bibitem{im2gps}
James Hays and Alexei~A. Efros,
\newblock ``im2gps: estimating geographic information from a single image,''
\newblock in {\em Proceedings of the {IEEE} Conf. on Computer Vision and
  Pattern Recognition ({CVPR})}, 2008.

\bibitem{geoMM}
{J. Choi, B. Thomee, G. Friedland, L. Cao, K. Ni, D. Borth, B. Elizalde, L.
  Gottlieb, C. Carrano, R. Pearce, D. Poland},
\newblock ``{ The Placing Task: A Large Scale Geo-Estimation Challenge for
  Social-Media Videos and Images},''
\newblock {\em 3rd ACM Multimedia Workshop On GeoTagging and Its Applications
  in Multimedia}.

\bibitem{huang07}
G.~B. Huang, M.~Ramesh, T.~Berg, and E.~Learned-Miller,
\newblock ``Labeled faces in the wild: A database for studying face recognition
  in unconstrained environments,'' .

\bibitem{deng09}
J.~Deng, W.~Dong, R.~Socher, L.~Li, K.~Li, and L.~Fei-fei,
\newblock ``Imagenet: A large-scale hierarchical image database,''
\newblock in {\em In CVPR}, 2009.

\end{thebibliography}
